\documentclass[a4paper, fleqn]{cas-dc}
\usepackage[numbers]{natbib} 

\makeatletter
\def\@evenfoot{}
\def\@oddfoot{}
\def\@footertext{}
\def\ps@first{%
  \def\@oddfoot{\leavevmode\hbox to \textwidth{\hfill\thepage\,/\,\pageref{LastPage}\hskip0.5cm}}%
  \def\@evenfoot{\leavevmode\hbox to \textwidth{\hfill\thepage\,/\,\pageref{LastPage}\hskip0.5cm}}%
}
\makeatother

\usepackage{fancyhdr}
\usepackage{lastpage}
\usepackage{graphicx} 
\usepackage{subfigure} 
\usepackage{algorithmic} 
\usepackage{algorithm} 
\usepackage{amsmath} 
\usepackage{multirow} 
\usepackage{multicol} 
\usepackage{tabularx}
\usepackage{orcidlink}

\begin{document}
 
\let\WriteBookmarks\relax
\def\floatpagepagefraction{1}
\def\textpagefraction{.001}

\shorttitle{Leveraging Label Semantics and Meta-Label Refinement for MLQC}
\shortauthors{Shi Dong et~al.}

\title [mode = title]{Leveraging Label Semantics and Meta-Label Refinement for Multi-Label Question Classification}

\author{Shi Dong}
\ead{dongshi@ccnu.edu.cn}

\author{Xiaobei Niu}[orcid=0009-0006-5265-7006]
\cormark[1]
\ead{nxb0302@mails.ccnu.edu.cn}
\cortext[corresponding]{Corresponding author.}

\author{Rui Zhong}
\ead{zhongr@ccnu.edu.cn}

\author{Zhifeng Wang}
\ead{zfwang@ccnu.edu.cn}

\author{Mingzhang Zuo}
\ead{mzzuo@ccnu.edu.cn}

\affiliation{
organization={Faculty of artificial intelligence in education ,Central China Normal University},
city={Wuhan},
state={Hubei},
country={China}
}

\begin{abstract}
Accurate annotation of educational resources is crucial for effective personalized learning and resource recommendation in online education. However, fine-grained knowledge labels often overlap or share similarities, making it difficult for existing multi-label classification methods to differentiate them. The label distribution imbalance due to the sparsity of human annotations further intensifies these challenges. To address these issues, this paper introduces RR2QC, a novel Retrieval Reranking method to multi-label Question Classification by leveraging label semantics and meta-label refinement. First, RR2QC improves the pre-training strategy by utilizing semantic relationships within and across label groups. Second, it introduces a class center learning task to align questions with label semantics during downstream training. Finally, this method decomposes labels into meta-labels and uses a meta-label classifier to rerank the retrieved label sequences.
In doing so, RR2QC enhances the understanding and prediction capability of long-tail labels by learning from meta-labels that frequently appear in other labels. Additionally, a mathematical LLM is used to generate solutions for questions, extracting latent information to further refine the model's insights. Experimental results show that RR2QC outperforms existing methods in Precision@K and F1 scores across multiple educational datasets, demonstrating its effectiveness for online education applications. The code and datasets are available at \url{https://github.com/78Erii/RR2QC}.
\end{abstract}

\begin{highlights}
\item Proposes RR2QC, a retrieval reranking method for classifying exercise content.
\item Utilizes a mathematical large language model for data augmentation.
\item Establishes a multi-label ranking contrastive learning pre-training task.
\item Addresses label imbalance and semantic overlap via class center and meta-label.
\end{highlights}


\maketitle

\section{Introduction}
\indent
In contemporary online learning environments, explicit annotation of knowledge labels is essential for transparency and interpretability in recommendation systems, supporting personalized teaching strategies. Nevertheless, the rapid growth of online education has resulted in an overwhelming volume of untagged exercises, making manual annotation costly, time-consuming, and prone to bias. This necessitates automatic exercise annotation methods to improve efficiency and reduce subjectivity in educational settings.

Exercise content, unlike news and product descriptions, typically includes descriptive text, mathematical formulas, and geometric images, structured in a question-solution format. Each exercise is assigned one or more knowledge labels, making automatic exercise annotation a Multi-Label Text Classification (MLTC) task. To learn richer semantic representations of exercises, traditional machine learning methods utilize techniques such as n-grams \cite{watt2008mathematical}, structural kernels \cite{suzuki2017mathematical}, and TF-IDF \cite{nguyen2012adaptive, huang2021automatic} to extract statistical information from mathematical symbols, combining these with text features obtained from word2vec embeddings. With the advancement of deep learning, feature-extracting encoders like TextCNN \cite{kim-2014-convolutional} and Bi-LSTM \cite{li2018news} have shown superior performance in processing complex text data. Pre-trained language models, particularly BERT \cite{devlin-etal-2019-bert}, are now widely used for text classification tasks. Certain works, such as MathBERT \cite{peng2021mathbert, Shen2021MathBERTAP}, JiuZhang \cite{zhao2022jiuzhang}, and QuesCo \cite{ning2023towards}, leverage specifically designed pre-training tasks in BERT to learn general representations of mathematical text. Furthermore, some researchers incorporate domain knowledge and knowledge graphs to extract additional features from exercise content, as seen in models like S-KMN \cite{S-KMN}, TGformer \cite{TGformer}, and SHGNet \cite{LI2023103348}. Although existing studies have achieved encouraging annotation results by enriching exercise representations, they do not address the unique challenges faced in the annotation process.

\begin{figure*}[t]
    \centering
    \includegraphics[width=0.9\textwidth]{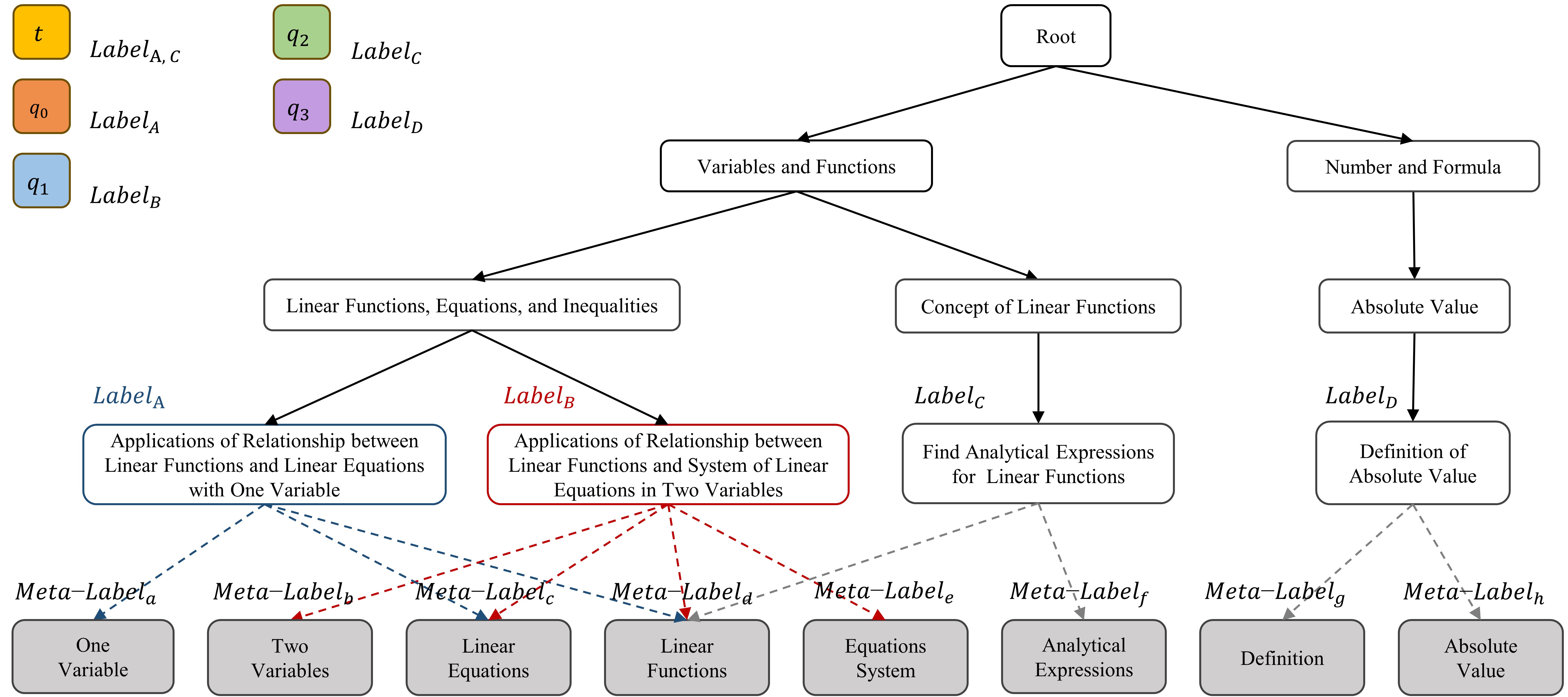}
    \caption{A four-level knowledge hierarchy tree and its meta-labels (bottom). As shown in the figure, $Label_A$ and $Label_B$ share the meta-labels "linear equations" and "linear functions," which easily mislead the classifier's decisions.}
    \label{knowledge tree}
\end{figure*}

In real-world online learning systems, automatic exercise annotation faces three key challenges. First, a small set of critical labels encompasses a large volume of exercises, while most other labels correspond to relatively fewer exercises, creating an uneven distribution that poses significant challenges for knowledge label prediction. Second, fine-grained long-text labels consist of multiple independent knowledge concepts (meta-labels), which can lead to semantic overlap in both label and exercise texts. For example, as shown in Figure \ref{knowledge tree}, both label A and label B contain meta-labels c and d, resulting in high similarity between their corresponding exercise content and making it difficult for classifiers to distinguish accurately. Lastly, some newer exercises lack reference solutions. Educational experts manually annotate these exercises by hypothesizing potential solution approaches, but models lack access to this prior knowledge. The absence of solutions further complicates the model’s comprehension of the exercises.

This paper introduces a Retrieval Reranking method to multi-label Question Classification (RR2QC) by leveraging label semantics and meta-label refinement, where "Question" specifically refers to exercises presented as mathematical text without solutions. First, RR2QC builds on QuesCo's framework to develop a foundational model for multi-label question understanding. QuesCo employs data augmentation and the knowledge hierarchy tree to create contrastive learning tasks that learn a holistic understanding of questions. Since QuesCo's pre-training tasks are designed for the single-label setting, we adapt them for the multi-label contexts. In detail, we introduce a ranking contrastive pre-training task that employs hierarchical knowledge distances to define positive sample pairs in multi-label contexts. By leveraging the semantic relationships among label groups, this approach leads to a more effective foundational model.

To address the decline in predictive performance due to semantic overlap and distribution imbalance of labels, the foundational model is divided into two independent models for downstream classification training. (1) The retrieval model predicts labels. To manage the extensive label set, we introduce a distance-based class center learning task that guides questions toward predefined class centers in the feature space. These class centers are learnable parameters derived from well-distributed label text vectors, enabling questions to focus on label textual information during training. The model then retrieves the label sequence most relevant to the question's semantics from a large pool of labels. (2) The reranking model leverages expert knowledge to decompose each label into meta-labels and retrains the foundational model on the meta-label dataset, generating a sequence of meta-labels for each question. By leveraging the mapping between labels and meta-labels, along with their confidence scores, the label sequence is reranked to produce a more refined result. This approach decomposes the semantically complex original labels into relatively independent and more evenly distributed meta-labels, aiding in the differentiation of highly similar labels and enhancing recognition of long-tail labels.

Furthermore, to address the issue of missing solutions for questions, we integrate answers generated by a mathematics large language model (Math LLM) into the question inputs to enrich their semantic content, filling gaps in data that traditional models might overlook.

Experiments demonstrate that RR2QC achieves superior Precision@k and F1 scores on real-world middle and high school mathematics and physics exercise datasets compared to existing text classification methods, establishing a new benchmark in the field. Additionally, RR2QC is applicable to other datasets with rich label semantics and uneven distributions. Our main contributions can be summarized as follows:
\begin{itemize}
\item[1)] We improve upon QuesCo by designing a ranking contrastive pre-training task suitable for the multi-label context.
\item[2)] We transform the text classification task into two steps: retrieval and reranking, effectively addressing the decline in automatic annotation accuracy caused by semantic overlap in lengthy text labels, as well as the prediction imbalance resulting from uneven label distribution.
\item[3)] We utilize the automatic problem-solving capabilities of a Math LLM to augment questions, effectively uncovering the latent semantics of exercises and enhancing classification performance.
\end{itemize}
The remainder of this paper is organized as follows: Section \ref{Related Work} provides a brief overview and discussion of related work. Section \ref{Method} presents our proposed RR2QC, including the necessity of using LLMs for data augmentation, the design of ranking contrastive pre-training task, and how to refine the retrieved label sequence by using meta-labels in downstream tasks. Section \ref{Experiment} reports on the experimental design and results analysis. Finally, Section \ref{Conclusion} concludes the paper.

\section{Related Work}\label{Related Work}
\subsection{Methods for class imbalance}
Due to the focus on education, exercises exhibit a long-tailed imbalanced distribution. Resampling and reweighting are the primary approaches currently employed to address class imbalance. Resampling simulates balanced data distributions by either downsampling or oversampling samples with high or low frequencies \cite{PouyanfarTMTKGD18, EstabrooksJJ04}. However, downsampling may result in the loss of key information, leading to underfitting. Conversely, oversampling can induce overfitting due to the repetition of low-frequency samples. Additionally, in multi-label scenarios, resampling may lead to over-sampling or under-sampling of common classes, limiting its applicability. On the other hand, reweighting, also known as cost-sensitive learning, rebalances classes by assigning greater weights to minority classes and smaller weights to majority classes during training. \citet{CuiJLSB19} introduces a class-balanced re-weighting term in the loss function, which is inversely proportional to the effective sample size for different classes. \citet{RidnikBZNFPZ21} designs an asymmetric loss that dynamically down-weights and hard-thresholds easy negative samples, ensuring a balanced contribution of positive and negative labels to the loss function. \citet{LiSWHCY24} proposes an adaptive distribution-based sample weighting method that considers inter-class imbalance ratios, intra-class density variables, and adaptive margins to address multi-class imbalance. \citet{ZhaoAXG24} treats label distribution as a continuous density function in the latent space, employing a flexible variational approach to approximate it in conjunction with the feature space, effectively addressing the class imbalance problem.

However, in educational datasets, labels exhibit both uneven distribution and strong correlations. For instance, the sample ratio of "Constructing Similar Triangles with Compass and Straightedge" to "Angle Calculation in Triangles" is starkly 1:100, yet both share the high-frequency meta-label "triangle." Unlike traditional resampling methods, we address this by decomposing original labels into independent meta-labels. This approach enhances the model's ability to recognize long-tail labels by learning a new, more balanced distribution of meta-labels without altering the sample distribution.

\subsection{Studies for Modeling Label Correlation}
Labels are not merely simple indices in classification; they carry specific semantics and exhibit strong correlations in multi-label contexts. Investigating these correlations and leveraging label semantics to enhance model performance has been a key focus of research. For instance, \citet{xiao2019label} converts label texts into vectors to measure semantic relationships with document words, constructing a label-specific document representation for training models. Similarly, \citet{guo2021label} learns label confusion to detect overlaps in label semantics by comparing text and label similarities, and generates an improved label distribution to replace the original one-hot label vector. Additionally, for datasets with hierarchical label distributions, such as WOS \cite{WOS} and RCV1-V2 \cite{RCV1}, recent studies in hierarchical multi-label text classification employ graph neural networks to learn label relationships or use supervised contrastive learning to enhance classification performance \cite{yu-etal-2023-instances, ZhangLSXTH24, wang-etal-2022-hpt, DPT}.

These works rely on datasets with clearly distinguishable and representative labels. In contrast, educational datasets involve multi-class classification with hundreds of labels, characterized by high semantic overlap. Previous algorithms find it challenging to identify key distinctions between labels. This paper introduces a distance-based class center learning task that adaptively enhances the proximity of similar samples, helping to define classification decision boundaries. Then, semantic overlap labels are decomposed into relatively independent meta-labels to refine the model's final predictions.
\subsection{LLM-based Text Classification}
The powerful linguistic capabilities of LLMs have spurred numerous innovations in text classification. \citet{sun-etal-2023-text} designs a clue and reasoning prompting method to guide GPT-3 in addressing the complex linguistic phenomena involved in text classification. \citet{li2023label} extracts the latent vectors from LLaMA to enhance LLM performance in tasks with limited labels. Drawing inspiration from knowledge distillation, \citet{hu2024llm} leverages GPT-3 to generate post analyses for social media posts, using these as augmented examples for contrastive learning, thereby enhancing the performance of the smaller model for personality detection. \citet{zhang2024teleclass} deploys LLMs to generate keywords for hierarchical categories, enhancing the feature and vector representations of the categories, and assisting in more refined document-category matching. Despite these advancements, the application of these methods to educational content remains challenging due to inconsistencies in data quality.
\section{Method}\label{Method}
\subsection{Problem Formulation}
\indent
The exercise dataset, denoted as $D = \{(q_i, Y_i)\}^{|D|}_{i=1}$, where $|D|$ is the total number of questions. Each question $q_i$ consists of descriptive text and mathematical formulas, and $Y_i = \{y_i\}^{|Y_i|}$ represents the label groups for $q_i$, with $|Y_i|$ indicating the number of labels in that group. The problem is to train a classification model that predicts top n labels $\hat{Y_i} = \{\hat{y_i}\}^n$ for each untagged question,  minimizing the discrepancy between $\hat{Y_i}$ and $Y_i$.

\subsection{Generating Solution by Math LLMs}
\begin{figure}[t]
    \centering
    \includegraphics[width=0.48\textwidth]{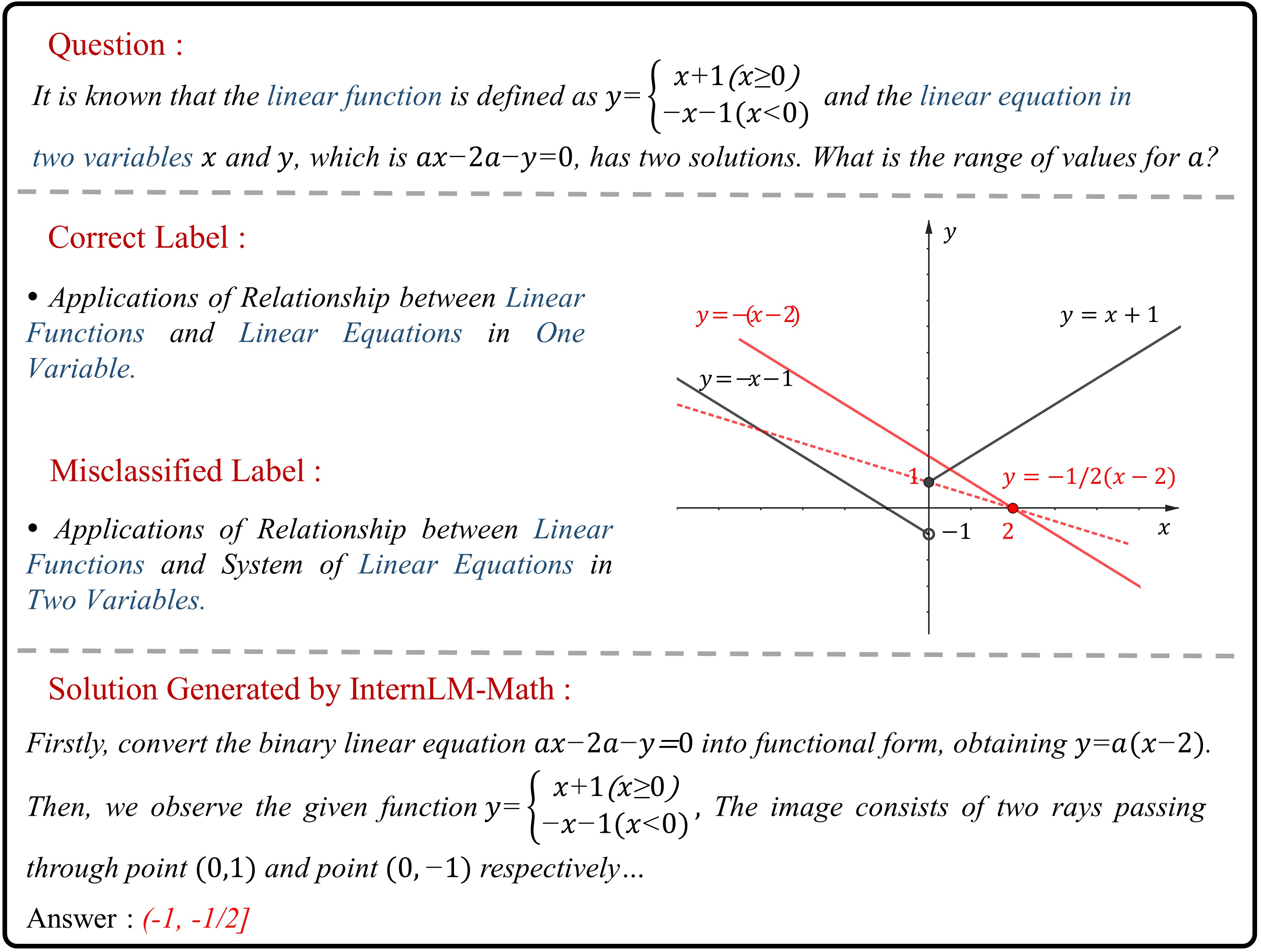}
    \caption{A question paired with a solution generated by the Math LLM. The blue phrases represent meta-labels.}
    \label{question_example}
\end{figure}

Conversational LLMs are generally not designed for tasks with fixed label outputs. Nevertheless, we exploit the mathematical reasoning capabilities of Math LLMs to improve the performance of smaller pre-trained models by generating detailed solutions for questions. For example, a question (see Figure \ref{question_example}) involves "Linear Equations in Two Variables" and "Linear Functions" in piecewise formats. This is frequently confused with "System of Linear Equations", mislabeling it as "Application of Relationship between Linear Functions and System of Linear Equations in Two Variables" instead of "Application of Relationship between Linear Functions and Linear Equations in One Variable". 

In the solution generated by a Math LLM, the conversion of a "Linear Equation in Two Variables" into a parameterized "Linear Equation in One Variable" clarifies the focus on intersection points between the linear function and equation. This approach offers an additional perspective on this exercise, which helps the classifier to distinguish between two highly similar labels. We evaluated prominent Math LLMs such as Llemma \cite{azerbayev2023llemma}, InternLM-Math \cite{ying2024internlmmath}, and DeepSeekMath \cite{deepseek-math}, all showing similar accuracy on exercise datasets. However, InternLM2-Math-Plus-20B, with its concise and standardized responses, was chosen to generate solutions, which were appended to the original question text. 

\subsection{Ranking Contrastive Pre-training}

\begin{figure*}[t]
    \centering
    \includegraphics[width=1.0\linewidth]{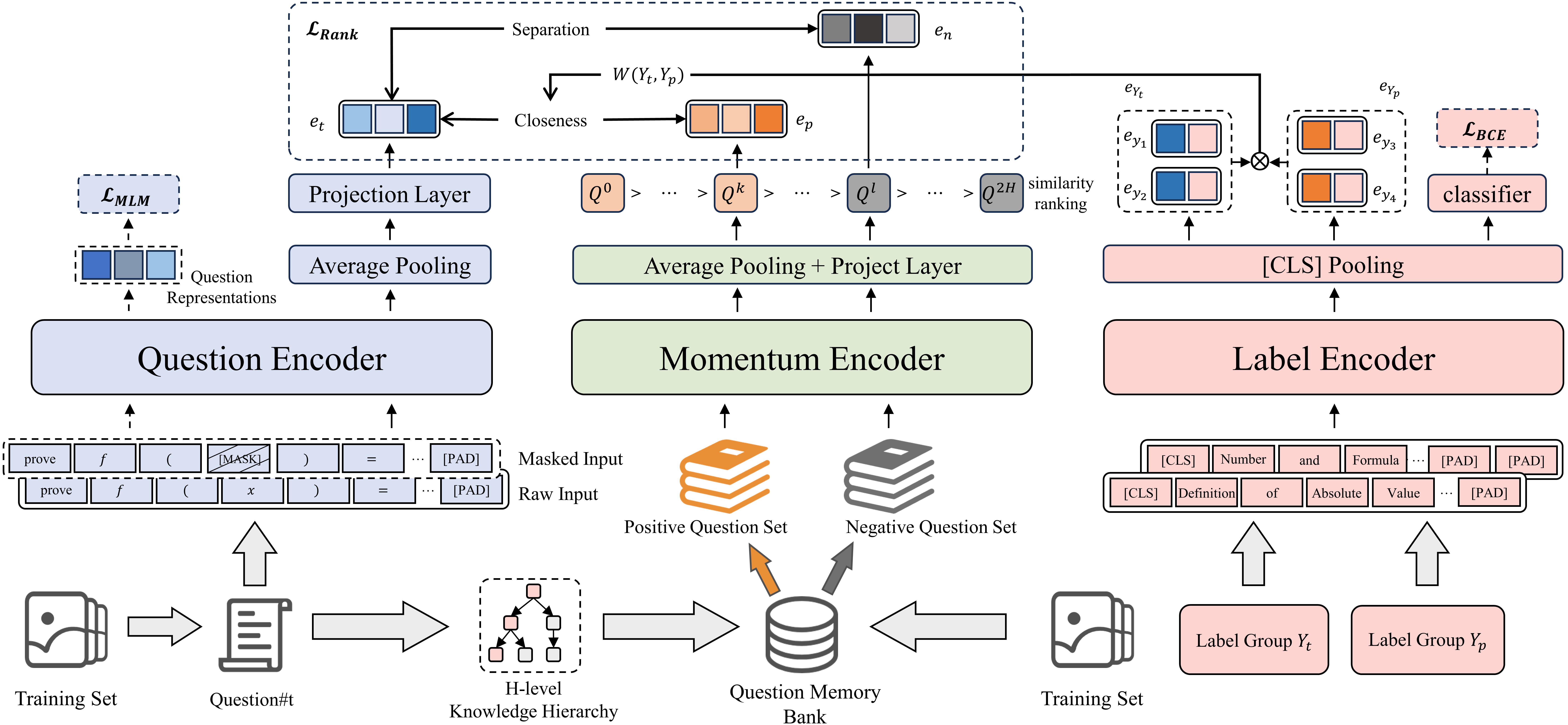}
    \caption{The general process of Ranking Contrastive Pre-training. The ranking contrastive learning between the left and the middle follows the MoCo framework,  while the label encoder on the right is used to train well-distributed label vectors to initialize the class centers in downstream task and simultaneously to weight the closeness between $e_t$ and $e_p$.}
    \label{pretrain}
\end{figure*}
\indent
The hierarchical structure of knowledge labels provides essential prior knowledge for guiding the model in measuring the relational distances between labels. QuesCo \cite{ning2023towards} utilizes this hierarchy to rank the similarities among questions and employs a MoCo-based \cite{he2020momentum} contrastive learning task to optimize its pre-training model. The MoCo framework in QuesCo employs a momentum encoder and a dynamic question queue to efficiently store and sample negative samples for contrastive learning, addressing the challenge of training with a large number of negatives. HJCL \cite{yu-etal-2023-instances} introduces a hierarchical penalty term to ensure the closeness of positive sample pairs within the same hierarchical level compared to those from more distant levels. However, these methods, implemented in single-label contexts, have not effectively addressed defining appropriate positive samples in multi-label contexts. To bridge this gap, we adapt QuesCo's knowledge hierarchy-aware rank module for multi-label settings during our pre-training phase, using a dynamic question queue to identify suitable positive samples $p$ for a target question $t$ at each hierarchical distance, controlling the closeness of samples based on their knowledge hierarchical distances from $t$.

Firstly, we define the distance between labels in a knowledge hierarchical tree. It is the sum of distances from the least common ancestor $lca(\cdot)$ to each label node:
\begin{equation}
    \begin{aligned}
        dis(y_i, y_j) = &depth(y_i) + depth(y_j) \\
        &- 2 \cdot depth(lca(y_i, y_j))\;,
    \end{aligned}
\end{equation}
based on this, the hierarchical distance between two questions $q_i$ and $q_j$ with label groups $Y_i$ and $Y_j$ can be defined as:
\begin{equation}
    khd({q_i}, {q_j}) = \left\{ {\begin{array}{*{20}{l}}
    0&\begin{array}{l}
    {\text{if}}\;\;\exists \;{y_i} \in {Y_i}, {y_j} \in {Y_j}, {\rm{ }}\\
    dis({y_i}, {y_j}) = 0
    \end{array}\\
    {\frac{{\sum dis({y_i}, {y_j})}}{{|{Y_i}| \cdot |{Y_j}|}}}&\begin{array}{l}
    {\text{if }} \;\;\forall \;{y_i} \in {Y_i}, {y_j} \in {Y_j}{\rm{,  }}\\
    dis({y_i}, {y_j}) > 0
    \end{array}
    \end{array}} \right.
    \label{khd}
\end{equation}
where $dis(\cdot) = 0$ indicates that two questions share at least one common label. For example, as shown in Figure \ref{knowledge tree}, question $t$ and $q_0$ share $Label_A$, resulting in $khd(t,q_0)=0$. Then, the set of all knowledge hierarchical distances can be represented as $khd(q_i, q_j)\in [0,\; 2H]$. $khd(\cdot) = 2H$ indicates the furthest distance, at which point it is unnecessary to pull the samples closer.

Secondly, given a target question $t$, our goal is to develop contrastive learning, which enables the model to accurately learn the fine-grained similarity rankings between questions with different distances, formalized as:
\begin{equation}
    sim(t, q_i) > sim(t, q_j)\;, \;\;\; \forall \; q_i \in Q^k,\; q_j \in Q^{l>k}\;,
\end{equation}
here, $sim(\cdot)$ is a similarity function computed as the dot product of question vectors, derived from the average of the last hidden states of an encoder after a projection layer. $Q^k$ denotes the set of questions at the same distance $k$ from $t$:
\begin{equation}
    Q^k = \{q_i \mid khd(t, q_i) = k\}\;, \quad k \in [0, \;2H].
\end{equation}

Thirdly, we adopt the Ranking Info Noise Contrastive Estimation (RINCE) loss \cite{hoffmann2022ranking} to optimize the model’s contrastive objective, defining it as follows:
\begin{equation}
    \mathcal{L}_\text{Rank} = \frac{1}{2H-1} \sum_{k=0}^{2H-1} \mathbf{l}_\text{k} \;,
    \label{L_Rank}
\end{equation}
where each rank loss $\mathbf{l}_\text{k}$ is defined as:
\begin{equation}
    \mathbf{l}_\text{k} = -\log \frac{\sum_{p \in Q^k} \exp\left(\frac{sim(t, p)}{\tau_k}\right) \cdot W(Y_t, Y_p)}{\cup_{l>k} \sum_{n \in Q^l} \exp \left(\frac{sim(t, n)}{\tau_l}\right)},
    \label{l_k}
\end{equation}
here, positive samples $p$ are pushed closer with the target $t$, while negative samples $n \in \cup_{l>k}Q^l$ are pushed farther away, with questions closer than $t$ having already optimized in $l_{khd(\cdot)<k}$. Besides, $\tau_k$ and $\tau_l$ are hyper-parameters for controlling the strength of penalties on positive and negative samples, typically set as $\tau_k < \tau_l$ to ensure dispersion of more distant negative samples in the feature space.

Subsequently, within the same rank distance, the similarity among label groups can vary significantly and should not be treated equally. The traditional Jaccard coefficient \cite{lin2023effective}, which simply calculates the intersection over the union of label groups, often overlooks the actual semantic relationships. As shown in Figure \ref{pretrain}, we weight the closeness of positive samples based on the dot product of label group vectors, considering both the overlap and the potential meanings. The weight function $W(\cdot)$ is calculated as follows:
\begin{equation}
    W({Y_t},{Y_p}) = \frac{1}{{|{Y_t}||{Y_p}|}}\sum\limits_{{y_i} \in {Y_t}} {\sum\limits_{{y_j} \in {Y_p}} {{\mathbf{e}_{{y_i}}} \cdot {\mathbf{e}_{{y_j}}}} }\;,
    \label{weight}
\end{equation}
here, $\mathbf{e}_{y_i}$ is the feature vector for label $y_i$.  When $\mathbf{e}_{y_i}$ is a one-hot vector, the Equation \ref{weight} is equivalent to the Jaccard coefficient. Additionally, to ensure that similar labels are closer in the feature space, we set up a classification task for leaf label texts, optimizing this task with Binary Cross-Entropy (BCE) loss:

\begin{equation}
    \begin{aligned}
        &\mathcal{L}_\text{BCE} = -\log{\mathbf{P}_i} \cdot \mathbf{y}^*_i + (1-\mathbf{y}^*_i) \cdot (1-\log{\mathbf{P}_i}), \\
        &\mathbf{P}_i = \frac{1}{1 + \exp{(-f_c(\mathbf{e}_{y_i}))}}\;, 
    \end{aligned}
    \label{BCE}
\end{equation}
where $f_c(\cdot)$ is a linear classifier and $\mathbf{y}^*_i$ is the multi-hot vector for label $y_i$.

Finally, we combine BERT's Masked Language Modeling (MLM), contrastive learning, and the classification task for leaf labels as our Ranking Contrastive Pre-training (RCPT) Task, jointly training the foundational model with these three losses:
\begin{equation}
    \mathcal{L}_\text{RCPT} = \mathcal{L}_\text{MLM} + \mathcal{L}_\text{Rank} + \mathcal{L}_\text{BCE}.
\end{equation}

\begin{algorithm}
\caption{The Ranking Contrastive Pre-training Algorithm}
    \label{RCPT algorithm}
    \begin{algorithmic}[1]
        \STATE \textbf{Inputs:}
        \STATE \quad $D$: Dataset of questions, $D = \{(q_i, Y_i)\}_{i=1}^{|D|}$, where $Y_i$ are label groups.
        \STATE \quad $K$: Dynamic question queue.
        \STATE \quad $\Theta_Q$: Question Encoder.
        \STATE \quad $\Theta_M$: Momentum Encoder.
        \STATE \quad $\Theta_L$: Label Encoder.
        \STATE \quad $m$: Momentum update parameter.
        \STATE \textbf{Outputs:}
        \STATE \quad Trained model parameters $\Theta_Q$ and $\Theta_L$.
        \STATE \quad Leaf label vectors $\{\mathbf{e}_{y_i}\}_{i=1}^n$.
        \STATE Initialize $\Theta_M \leftarrow \Theta_Q$.
        \STATE Define $dis(y_i, y_j)$ using the least common ancestor.
        \STATE \textbf{for} each epoch $e \in \{1, \ldots, n\}$ \textbf{do}
        \STATE \quad \textbf{for} each minibatch  $t \in \{1, \ldots, M\}$ \textbf{do}
        \STATE \quad \quad Calculate hierarchical distances $khd(q_t, q_j)$ for all $q_j \in K$.
        \STATE \quad \quad Update $\Theta_Q$ using:
        \STATE \quad \quad \quad $\Theta_Q \gets \Theta_Q - \alpha \nabla_{\Theta_Q} (\mathcal{L}_\text{Rank}+\mathcal{L}_\text{MLM})$
        \STATE \quad \quad Update $\Theta_M$ using:
        \STATE \quad \quad \quad $\Theta_M \gets m \cdot \Theta_M + (1-m) \cdot \Theta_Q$
        \STATE \quad \quad Update $\Theta_L$ using:
        \STATE \quad \quad \quad $\Theta_L \gets \Theta_L - \beta \nabla_{\Theta_L} \mathcal{L}_\text{BCE}$
        \STATE \quad \textbf{end for}
        \STATE \quad Manage the question queue: enqueue and dequeue.
        \STATE \textbf{end for}
        \STATE Calculate leaf label vectors $\{\mathbf{e}_{y_i}\}_{i=1}^n$ from $\Theta_L$.
    \end{algorithmic}
\end{algorithm}

The general process and algorithm of RCPT can be referenced in Figure \ref{pretrain} and Algorithm \ref{RCPT algorithm}.

\subsection{Retrieval Reranking To Question Classification}

In the downstream phase, the pre-trained foundational model is split into two distinct models: a retrieval model that retrieves the most relevant label sequences and a reranking model refining these sequences, addressing the semantic overlap and distribution imbalance of labels.

\subsubsection{Class Center Learning.} 
First, instead of directly applying $\mathcal{L}_\text{BCE}$ for classification, we introduce a distance-based Class Center Learning (CCL) task during the training of the retrieval model. CCL pulls questions toward their corresponding predefined class centers, reducing ambiguous samples along the decision boundary and enhancing the classification space partitioning. 

This task is an adaptation of the Class Anchor Clustering (CAC) method proposed by \citet{CAC}. CAC is used in open set recognition to train known classes to form tight clusters around anchored class centers in the feature space. \citet{PAMI} treats class centers as learnable parameters, initializing them randomly while focusing on enhancing intra-class compactness and maintaining inter-class separation.

In contrast to previous methods, CCL represents each class center $ \mathbf{c}_{y_i} $ as a learnable parameter initialized from feature vectors $ \mathbf{e}_{y_i} $ (see Step 25 of Algorithm \ref{RCPT algorithm}). These well-distributed pre-trained $ \mathbf{e}_{y_i} $, derived from label texts, naturally ensure that semantically similar labels have closer class centers, avoiding issues caused by random initialization. Unlike \citet{PAMI}, CCL avoids explicit inter-class distancing to preserve the semantic relationships embedded in the class centers. We assist the retrieval model's training by minimizing the Euclidean distance between the question's feature vector and its corresponding class center, defined as:
\begin{equation}
    \mathcal{L}_\text{CCL} = \frac{1}{M}\sum_{i=1}^M \parallel\; \mathbf{e}_{q_i} - \mathbf{c}_{Y_i} \parallel^2_2\;, 
\end{equation}
here, $\mathbf{e}_{q_i}$ is the feature vector of $q_i$, and $\mathbf{c}_{Y_i}$ is the mean of the class centers for $q_i$'s multi-labels, with $M$ as the minibatch size. We optimize $\mathcal{L}_\text{BCE}$ (Equation \ref{BCE}) for accurate classification, adjusting $\lambda$ to balance the influence of CCL, and jointly train the retrieval model using both losses:
\begin{equation}
    \mathcal{L}_\text{Retrieval} = \mathcal{L}_\text{BCE} + \lambda \mathcal{L}_\text{CCL}\;.
    \label{Retrieval}
\end{equation}

\begin{figure*}
    \includegraphics[width=1.\linewidth]{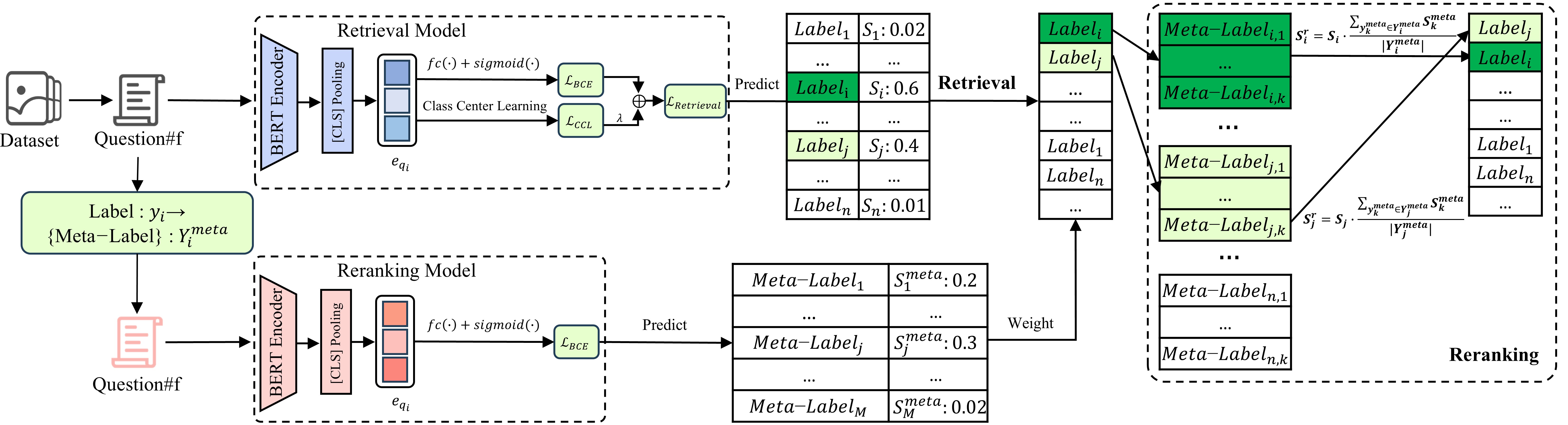}
    \caption{The overall process of Retrieval Reranking, where the retrieval model generates initial label predictions, which the reranking model refines using meta-labels. Final scores are calculated by combining retrieval label scores with weighted meta-label scores to enhance ranking performance.}
    \label{retrieval reranking}
\end{figure*}

The retrieval model outputs a label sequence for an untagged question $f$ with confidence, formally represented as:

\begin{equation}
    Y_f = \{y_i:s_i\}^N_{i=1},\quad s_i \in (0, 1)\;, 
    \label{label}
\end{equation}
where $s$ is the confidence score after passing the question features through a classifier and an activation function. $N$ denotes the total number of labels. Next, as shown in Figure \ref{retrieval reranking}, the reranking model further refines the label sequence by distinguishing labels that the retrieval model finds difficult to differentiate.

\subsubsection{Meta-label Refinement.}\label{manual decomposition}
Second, in educational datasets, long-text labels often combine multiple knowledge concepts, referred to as meta-labels, which correspond to different parts of the question content. This is the fundamental cause of semantic overlap, and extracting meta-labels is the most direct solution. Building on this, with the support of a meta-label classifier (reranking model), the confidence scores of meta-labels guide the retrieval model by highlighting the knowledge concept most relevant to the question, thereby effectively improving the ranking of ground truth within the retrieval label sequence.

While existing approaches\cite{metalabel2014, metalabel2016, metalabel2021} construct meta-labels through clustering or random strategies, manually decomposing meta-labels in educational datasets offers a simpler approach with broader semantic coverage. This shifts question labeling from emphasizing concept intersections to embracing their union. Specifically, we engage experts to manually decompose each label $y_i$ into meta-labels $Y^{\text{meta}}_i$, constructing a mapping relationship from labels to meta-labels:

\begin{equation}
    \{y_i\to Y^{meta}_i\}^N_{i=1}, \quad Y^{meta}_i \in Y^{meta} = \{y^{meta}_j\}^M_{j=1}\;.
    \label{label2metalabel}
\end{equation}

\begin{table}[t]
    \scalebox{0.95}{
    \centering
    \begin{tabularx}{0.5\textwidth}{p{4.5cm}|p{3.5cm}}
        \hline
        \textbf{Label (question numbers)} & \textbf{Meta-labels} \\
        \hline
        Angle Calculation in Triangles \newline(876) & Triangle, Calculation,\newline Geometric Angles \\
        \hline
        Using Pythagorean Theorem for Calculation (499)& Pythagorean Theorem,\newline Calculation \\
        \hline
        Practical Applications of Polynomial Addition and Subtraction (209)& Polynomial, Four Arithmetic Operations, Practical Application \\
        \hline
        Setting Up Algebraic Expressions for Real-World Problems (158)& Algebraic Expression,\newline Practical Application \\
        \hline
        Applications of Inequality Properties (128)& Inequality Analysis,\newline Optimization Techniques \\
        \hline
        Pythagorean Triples Problem (37)& Pythagorean Theorem,\newline Triples \\
        \hline
        Constructing Similar Triangles\newline with Compass and Straightedge (8)&Compass and Straightedge, Similar Triangles, Construction, Triangle \\
        \hline
        Solving Compound Inequalities \newline(7)& Compound Inequalities,\newline Inequality Analysis \\
        \hline
    \end{tabularx}}
    \caption{Some examples of the label-to-meta-label mapping dictionary, manually decomposed by experts.}
    \label{label2meta-label_dict}
\end{table}

Referring to Table \ref{label2meta-label_dict}, the steps and principles for experts to decompose meta-labels are as follows:
\begin{itemize}
\item[1)] Extract independent phrases from the labels to form candidate meta-labels.
\item[2)] Experts restrict synonyms among meta-labels to a single usage based on the definitions of knowledge labels within the educational context.
\item[3)] Ensure that each meta-label appears in multiple original labels.
\item[4)] Since candidate meta-labels may not be entirely semantically independent, subordinate meta-labels with hierarchical relationships should include their parent meta-labels. For instance, "Isosceles Triangle" and "Triangle" are related; we incorporate "Triangle" into the decomposition of "Understanding and Applying Isosceles Triangles" to indicate that these questions encompass both concepts.
\item[5)] Common meta-labels that appear in more than 10\% of the labels, such as "Relationship" and "Application," which cannot effectively distinguish text content, should be removed from the candidate to reduce the output space of the reranking model.
\end{itemize}

Regarding whether expert-driven meta-label decomposition might limit its real-world application, we automate this process using a LLM and compare the results with  those of expert decompositions. The experimental design and findings are detailed in section \ref{Comparison between manual and automatic decomposition}, highlighting the significance of expert knowledge and the quality of meta-labels.

Subsequently, we transform the original dataset $D$ into a meta-label dataset $D^{meta} = \{q_i, Y^{meta}_i\}^{|D^{meta}|}_{i=1}$, using $\mathcal{L}_\text{BCE}$ (Equation \ref{BCE}) to train the reranking model for classifying these meta-labels. As outlined in Equation \ref{label}, the reranking model outputs a meta-label sequence for question $f$:

\begin{equation}
    Y^{\text{meta}}_f = \{y^{\text{meta}}_j: s^{\text{meta}}_j\}^M_{j=1}\;, \quad s^{\text{meta}}_j \in (0,1)\;.
    \label{metalabel}
\end{equation}

Based on Equations \ref{label}, \ref{label2metalabel}, and \ref{metalabel}, we refine the retrieval label sequence of question $f$ as follows:

\begin{equation}
    \begin{aligned}
        &Y^{r}_f = \{y_i: s^r_i\}^N_{i=1}\;,\quad s^r_i \in (0,1)\;,\\
        &s^r_i = s_i \cdot \frac{1}{|Y^{\text{meta}}_i|}\sum_{y^{\text{meta}}_j \in Y^{\text{meta}}_i} s^{\text{meta}}_j\;.
    \end{aligned}
    \label{refine}
\end{equation}

The reorganization and weighting of meta-labels enable RR2QC to make more accurate classification decisions among similar labels. When the model is uncertain between labels with similar confidence levels, RR2QC uses the confidence of the corresponding meta-labels to better identify which knowledge concepts in the question are most critical. Moreover, long-tail labels decomposed into high-frequency and long-tail meta-labels are easier for the model to learn, enhancing their ranking within the label sequence. Ultimately, we reorder the label sequence $Y^{r}_f$ based on confidence score $s^r$ and select the top $n$ labels as the predicted label set $\hat{Y} = \{\hat{y}\}^n$ for question $f$.

\begin{table}[t]
    \scalebox{0.9}{
    \centering
    \begin{tabular}{c|cccc}
        \hline
         Dataset&MJ&MS&PS&DA-20K/$\text{DA-20K}_\text{AUG}$\\
         \hline
         D&45610&132297&44057&22417/22417\\
         H&4&3&3&1/1\\
         L&721&636&384&427/427\\
         M&365&274&298&289/289 \\
         $\overline{\text{T}}$&163.39&212.46&353.31&134.85/283.46\\
         $\overline{\text{L}}$&1.11&1.42&1.26&1.76/1.76\\
         $\overline{\text{M}}$&2.86&3.89&3.57&4.40/4.40 \\
         \hline
    \end{tabular}}
    \caption{Statistics of Math Junior(MJ),  Math Senior(MS),  Physics Senior(PS),  DA-20K and the augmented DA-20K datasets. D and H denote the total number of questions and hierarchical levels. L and M denote the number of labels and meta-labels. $\overline{\text{T}}$ denotes average token per question. $\overline{\text{L}}$ and $\overline{\text{M}}$ denote average number of labels and meta-labels per question.}
    \label{datasets}
\end{table}

\newpage
\section{Experiment}\label{Experiment}
\subsection{Datasets}
We conduct experiments on four datasets: Math Junior, Math Senior, Physics Senior, and the publicly available DA-20K dataset \cite{wang2022automatic}. Each dataset, sourced from Chinese educational resources and exam papers, is annotated by professional educators. The datasets are randomly split into training, validation, and testing sets in an 8:1:1 ratio. Mathematical expressions with \LaTeX{} format in the question content are tokenized by the im2markup tool \cite{deng2017image}. For DA-20K, we generate solutions for a random one-third of the questions using InternLM2-Math-Plus-20B and concatenate these with the questions to create the augmented dataset $\text{DA-20K}_\text{AUG}$. Statistical information for all datasets is provided in Table \ref{datasets}.

\subsection{Baseline methods}
Due to the scarcity of algorithms for automatic question annotation, we compare our proposed method with the following state-of-the-art text classification methods.
\begin{itemize}
    \item \textbf{TextCNN \cite{kim-2014-convolutional}}: The first model to introduce convolutional neural networks into text classification.  
    \item \textbf{AttentionXML \cite{you2019attentionxml}}: This LSTM-based model for large-scale multi-label classification integrates a label tree with an attention mechanism to focus on relevant labels, reducing the candidate set and improving both classification efficiency and accuracy, while also addressing label imbalance.
    \item \textbf{Vanilla BERT/RoBERTa}: These models are applied directly for classification tasks without pre-training, using BERT-Base-Chinese \cite{devlin-etal-2019-bert} and RoBERTa-zh-Large \cite{liu2019roberta}.
    \item \textbf{TAPT BERT/RoBERTa}: Similar to Vanilla but with Task-Adaptive Pre-Training (TAPT) \cite{gururangan-etal-2020-dont} on the target dataset using the MLM task.
    \item \textbf{QuesCo (BERT)\cite{ning2023towards}}: A ranking contrastive pre-training model that integrates a single-label hierarchy and data augmentation, subsequently freezing the encoder layers for classifier training. We apply its pre-training method only to single-label training datasets.
    \item \textbf{ChatGLM3} \cite{glm2024chatglm}: An open-source large language model fine-tuned through parameter-efficient techniques (e.g., LoRA \cite{hu2021lora}), which we use ChatGLM3-6B and primarily compare its performance through Precision at top one label, given the variable sizes of the label sets it generates.
    \item \textbf{Qwen2.5} \cite{Yang2024Qwen25TR}: An open-source LLM series by Alibaba Cloud (0.5B–72B parameters) that excels in instruction following, long-text generation, and structured data processing. We evaluate Qwen2.5-7B-Instruct and compare its P@1 score under the same setup as ChatGLM3-6B.
    \item \textbf{HPT (BERT)} \cite{wang-etal-2022-hpt}: A Hierarchy-aware Prompt Tuning method for hierarchical text classification (HTC) from a multi-label MLM perspective. HPT fuses label hierarchy knowledge through dynamic virtual templates and soft prompts, and introduces a zero-bounded multi-label cross-entropy loss to align HTC and MLM objectives.

    \item \textbf{HALB (BERT)} \cite{ZhangLSXTH24}: This Hierarchical-Aware and Label-Balanced model for hierarchical multi-label text classification, uses a graph neural network for label features, an attention mechanism for hierarchical-aware representations, and asymmetric loss to address label imbalance.
    
    \item \textbf{DPT (BERT)} \cite{DPT}: This paper uses Prompt Tuning and contrastive learning to highlight label discrimination across hierarchical layers. It introduces a hand-crafted prompt with slots for positive and negative label predictions, and a label hierarchy self-sensing task to ensure cross-layer consistency.
\end{itemize}

\subsection{Experimental environment}
BERT-Base-Chinese is chosen as the base model for RR2QC. In the ranking contrastive pre-training task, we set the projection layer's output dimension to 128 and use a momentum encoder update parameter $m$ of 0.9. The queue size is 1\% of the training set size, with temperature factors $\tau$ for $L_{Rank}$ adjusted geometrically from 0.1 to 0.6. For the downstream task, following standard setups \cite{ingle2022investigating}, we freeze half of the encoder layers, maintain a learning rate of 2e-5, use a batch size of 32, set $\lambda$ to 0.1, and the number of epochs to 30. Additionally, we employ two graduate students majoring in education for the manual decomposition of meta-labels. All experiments are conducted on a computer equipped with an NVIDIA GeForce RTX 4090 GPU.

\subsection{Evaluation metrics}
To evaluate the performance of RR2QC and these baselines on imbalanced multi-label datasets, we select Precision at top K (P@K), Macro-F1, and Micro-F1 as evaluation metrics. P@K intuitively calculates the number of correct labels within the top K predicted labels. Micro-F1 is more sensitive to high-frequency classes, while Macro-F1 is more sensitive to low-frequency classes. These metrics are defined as follows:
\begin{equation}
    \begin{aligned}
        &Micro-F1 = \frac{2 \times  P\text{@}1 \times R\text{@}1}{P\text{@}1 + R\text{@}1}\;, \\
        &\text{where}\; P\text{@}K = \frac{\sum TP\text{@}K}{\sum TP\text{@}K+\sum FP\text{@}K}\;, \\
        &R\text{@}K = \frac{\sum TP\text{@}K}{\sum TP\text{@}K+\sum FN\text{@}K}\;,
    \end{aligned}
\end{equation}
\begin{equation}
    \begin{aligned}
        &Macro-F1 = \frac{1}{N}\sum_{i=1}^N\frac{2 \times P_i \times R_i}{P_i + R_i}\;, \\
        &\text{where}\; P_i = \frac{TP_i}{TP_i + FP_i}\;, R_i = \frac{TP_i}{TP_i + FN_i}\;,
    \end{aligned}
\end{equation}
here $N$ denotes the number of labels, while $TP$, $FP$, and $FN$ represent true positives, false positives, and false negatives, respectively. The index $i$ refers to the $i$-th label, and $\text{@}K$ indicates the top $K$ labels based on their confidence scores.

\subsection{Experimental results}

\begin{table*}[h]
    \centering
    \scalebox{0.85}{
    \begin{tabular}{l|cccc|cccc|cccc}
        \hline
         \multirow{2}{*}{Model}&\multicolumn{4}{c}{Math Junior}&\multicolumn{4}{c}{Math Senior}&\multicolumn{4}{c}{Physics Senior}\\
         \cline{2-13}
         &P@1&P@2&Micro-F1&Macro-F1&P@1&P@2&Micro-F1&Macro-F1&P@1&P@2&Micro-F1&Macro-F1\\
         \hline
         \hline
         TextCNN&74.27&44.73&70.55&58.08&81.49&55.09&67.19&51.77&76.11&49.77&67.55&55.23\\
         AttentionXML&71.70&43.18&68.11&55.19&81.49&55.18&67.19&51.61&75.47&49.47&66.99&55.52\\
         $\text{BERT}_\text{Base}$&74.29&44.36&70.56&56.52&83.25&55.78&68.63&\underline{55.04}&76.67&50.05&68.06&56.30\\
         $\text{RoBERTa}_\text{Large}$&74.25&44.33&70.53&58.39&83.06&55.77&68.47&53.86&76.61&49.35&68.00&54.03\\
         BERT-TAPT&75.36&44.83&71.58&59.49&\underline{84.21}&56.47&\underline{69.43}&53.92&77.31&50.26&68.62&54.44\\
         RoBERTa-TAPT&75.17&44.51&71.40&58.05&83.65&\underline{56.50}&68.97&53.63&\underline{78.21}&50.61&\underline{69.42}&53.73\\
         QuesCo&76.05&43.88&72.24&\underline{62.35}&82.01&54.69&67.61&52.48&77.05&49.92&68.39&\underline{56.40}\\
         ChatGLM3&75.03&-&70.56&59.57&78.51&-&64.49&49.23&73.28&-&64.86&52.88\\
         Qwen2.5&73.68&-&69.98&60.75&80.29&-&66.19&54.19&75.32&-&66.85&55.73\\
         HALB&75.29&44.45&71.51&59.38&82.22&55.79&67.78&49.91&76.63&49.91&68.02&54.27\\
         HPT&\underline{76.42}&\textbf{45.41}&\underline{72.58}&60.99&83.85&\textbf{56.53}&69.13&55.01 &76.68&50.60&68.06&56.38\\
         DPT&75.75&45.10&71.94&62.17&83.62&56.32&68.94&54.51 &78.12&\underline{50.81}&69.34&55.97\\
         \hline
         \hline         
         RR2QC&\textbf{78.36}&\underline{45.15}&\textbf{74.42}&\textbf{64.03}&\textbf{84.92}&55.65&\textbf{70.00}&\textbf{55.14}&\textbf{79.68}&\textbf{50.93}&\textbf{70.72}&\textbf{56.60}\\
         \hline
    \end{tabular}}
    \caption{Results of RR2QC and baseline methods on Math Junior, Math Senior and Physics Senior dataset. The best results are highlighted in bold, while the second-best are underlined.}
    \label{MJ MS PS}
\end{table*}
\begin{table}[h]
    \centering
    \scalebox{0.75}{
    \begin{tabular}{l|cccc}
        \hline
         \multirow{2}{*}{Model}&\multicolumn{4}{c}{DA-20K/$\text{DA-20K}_\text{AUG}$}\\
         \cline{2-5}
         &P@1&P@2&Micro-F1&Macro-F1\\
         \hline
         \hline
         TextCNN&60.90/61.70&47.35/45.92&44.22/44.80&32.15/32.13\\  
         AttentionXML&60.14/60.23&46.50/46.79&43.67/43.74&30.95/30.91\\
         $\text{BERT}_\text{Base}$&60.77/60.72&45.76/45.89&44.12/44.09&31.61/31.87\\
         $\text{RoBERTa}_\text{Large}$&60.19/60.49&45.99/46.50&43.93/43.74&30.85/30.84\\
         BERT-TAPT&60.49/61.39&46.03/46.90&43.93/44.58&32.12/32.85\\
         RoBERTa-TAPT&\underline{61.48}/\underline{62.33}&\textbf{46.88}/\underline{47.33}&\underline{44.64}/\underline{45.26}&\underline{32.33}/\underline{32.79}\\
         
         ChatGLM3&55.51/55.76&-&40.08/40.27&31.90/31.27\\
         Qwen2.5&56.26/58.27&-&40.85/42.31&31.54/32.61\\
         \hline
         \hline         RR2QC&\textbf{62.86}/\textbf{64.16}&\underline{46.83}/\textbf{47.49}&\textbf{45.65}/\textbf{46.58}&\textbf{35.12}/\textbf{34.11}\\
         \hline
    \end{tabular}}
    \caption{Results of RR2QC and baseline methods on DA-20K and $\text{DA-20K}_\text{AUG}$ dataset. The best results are highlighted in bold, while the second-best are underlined.}
    \label{DA-20K}
\end{table}

To validate the performance of our proposed RR2QC in multi-label question classification tasks, we conduct extensive experiments across four educational datasets, comparing RR2QC against 12 typical baseline methods. We evaluate performance using four main metrics: P@1, P@2, Micro-F1, and Macro-F1. As shown in Tables \ref{MJ MS PS} and \ref{DA-20K}, RR2QC achieves the best results in P@1, Micro-F1, and Macro-F1 across all datasets, demonstrating its effectiveness.

In the Math Junior, Math Senior and Physics Senior datasets, RR2QC's P@1 surpasses the second-best model by 1.94\%, 0.71\% and 1.47\%, respectively, showcasing RR2QC's accuracy in predicting the most relevant labels. For P@2, RR2QC performs comparably to the second-best model, indicating robust accuracy in identifying the top two labels. Micro-F1 improvements of 1.84\%, 0.57\% and 1.30\% demonstrate RR2QC's strong overall prediction performance, as Micro-F1 measures prediction accuracy across all labels, revealing RR2QC’s consistency across samples and labels. Macro-F1 gains of 1.68\%, 0.10\%, and 0.20\% highlight RR2QC’s ability to address long-tail labels, achieving high recall. 

On DA-20K and its augmented dataset $\text{DA-20K}_\text{AUG}$, RR2QC improves P@1, Micro-F1, and Macro-F1 by at least 1.38\%, 1.01\%, and 2.79\% over the second-best model, demonstrating its adaptability to large-scale datasets and long-tail label challenges. Across all models, performance improves on $\text{DA-20K}_\text{AUG}$, confirming the effectiveness of Math LLM-based data augmentation. The solutions generated by Math LLM add richer context, enhancing the model’s understanding of exercise semantics and multi-label classification performance. The HTC models (QuesCo, HPT, HALB, DPT) were not evaluated due to DA-20K’s single-label layer, which is incompatible with the hierarchical design of HTC models.

Across all datasets, QuesCo and TAPT BERT/RoBERTa (models enhanced through further pre-training based on BERT) achieve strong results, highlighting the importance of pre-training tasks and BERT's effectiveness in MLTC. Surprisingly, AttentionXML, intended for large-scale multi-label datasets and long-tail labels, performs below the CNN-based TextCNN and Transformer-based models (e.g., BERT, RoBERTa). This may be due to AttentionXML's use of an LSTM encoder, which is less effective than BERT at capturing long-range information and less efficient than CNN at extracting local features. ChatGLM3 and Qwen2.5 show the weakest performance, suggesting that conversational LLMs are generally unsuitable for classification tasks with a fixed output space. Meanwhile, HALB, HPT, and HPT's optimized variant, DPT, are three HTC models. In terms of performance, HALB performs the worst, while HPT and DPT show better results, particularly achieving the best and second-best P@2 scores. This could be due to the use of the prompt-tuning paradigm in HPT and DPT, which combines pre-training tasks with classification. In contrast, HALB assumes labels follow a hierarchical tree structure, utilizing graph convolutional networks to compute hierarchical label features and filter text representations accordingly. However, while educational datasets organize knowledge labels hierarchically, predictions focus only on the leaf nodes as relevant labels, which somewhat limits HALB’s ability to model label relevance effectively.

\begin{table*}[t]
\scalebox{0.80}{
    \centering
    \begin{tabular}{l|ccc|ccc|ccc}
        \hline
         \multirow{2}{*}{Model}&\multicolumn{3}{c|}{head($\ge 72$)}&\multicolumn{3}{c|}{medium(26-72)}&\multicolumn{3}{c}{tail($\le 26$)}\\
         \cline{2-10}
         &P@1&Micro-F1&Macro-F1&P@1&Micro-F1&Macro-F1&P@1&Micro-F1&Macro-F1\\
         \hline\hline
         TextCNN&79.16&71.14&65.37&64.18&57.35&65.23&42.47&38.82&64.26\\
         AttentionXML&76.23&68.50&62.48&62.57&55.91&62.43&41.40&37.84&61.63\\
         $\text{BERT}_\text{Base}$&79.73&71.65&\underline{69.58}&65.22&58.28&66.56&40.05&36.61&62.18\\
         $\text{RoBERTa}_\text{Large}$&78.65&70.68&67.55&67.11&59.97&\textbf{71.02}&46.24&42.26&66.76\\
         BERT-TAPT&79.85&71.76&65.47&67.77&60.56&67.96&46.51&42.51&65.24\\
         RoBERTa-TAPT&79.99&71.88&\textbf{71.22}&67.86&60.64&\underline{70.14}&42.74&39.07&64.43\\
         QuesCo&79.29&71.26&65.06&\underline{70.89}&\underline{63.34}&68.42&51.35&46.43&68.25\\
         ChatGLM3&77.99&69.54&64.40&66.47&58.49&64.79&51.48&45.49&67.50\\
         Qwen2.5&76.74&68.96&64.25&66.54&59.46&65.85&48.92&44.72&67.05\\
         HALB&79.76&71.67&63.71&67.01&59.88&65.56&49.73&45.45&66.73\\
         HPT&79.02&71.01&64.40&69.66&62.25&67.71&\underline{51.88}&\underline{47.42}&\underline{68.41}\\
         DPT&\underline{80.82}&\underline{72.63}&66.44&69.57&62.16&67.48&49.46&45.21&67.49\\
         \hline\hline         RR2QC&\textbf{81.81}&\textbf{73.52}&65.92&\textbf{72.50}&\textbf{64.78}&68.56&\textbf{53.23}&\textbf{48.65}&\textbf{69.04}\\
         \hline
    \end{tabular}}
    \caption{Results of our method and other baselines for the head, middle, and tail subsets of labels in Math Junior dataset. The best results are highlighted in bold, while the second-best are underlined.}
    \label{imbanlance test}
\end{table*}

\subsection{Model analysis}
\subsubsection{Ablation Study}
\begin{figure}
    \centering
    \includegraphics[width=1.0\linewidth]{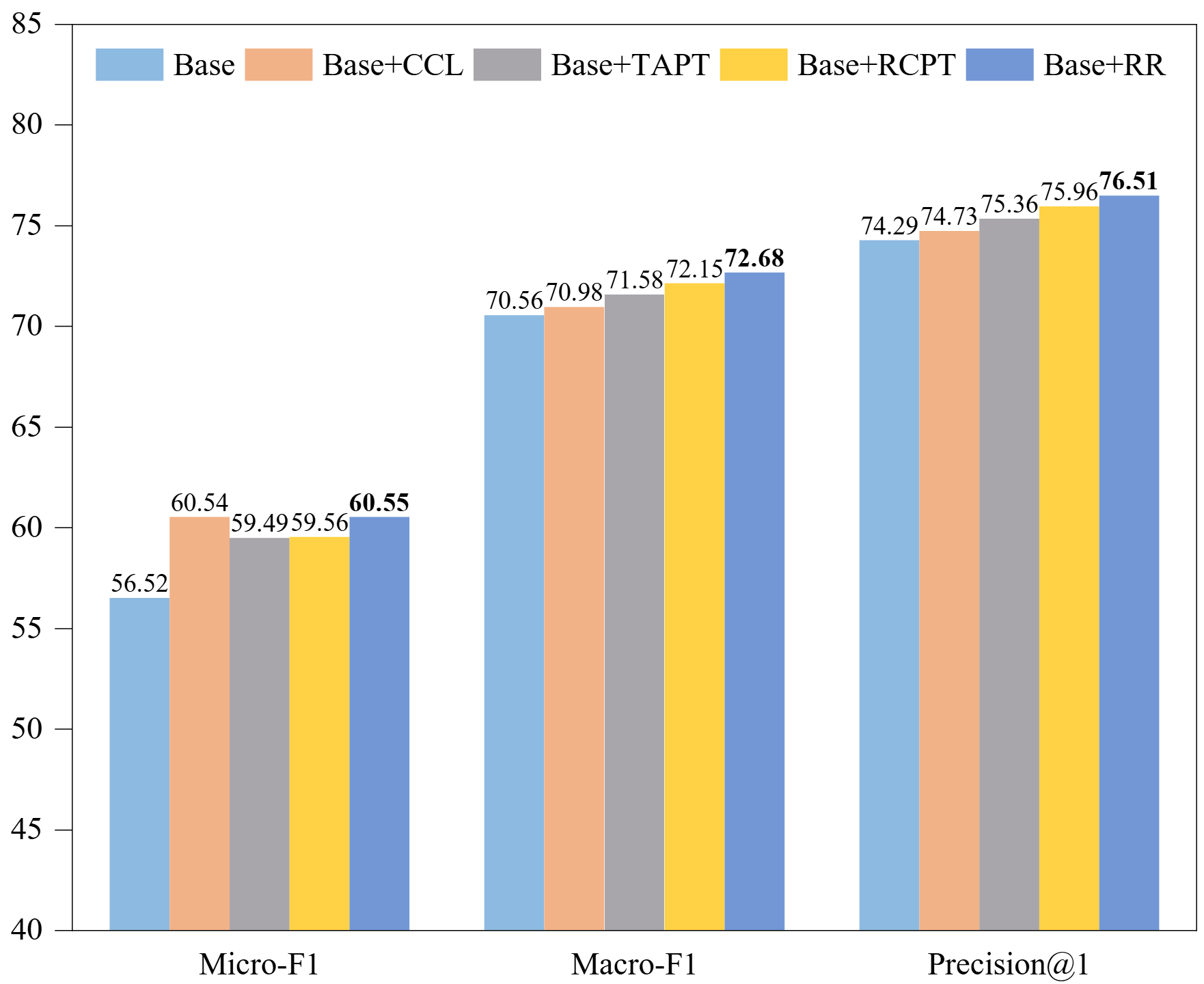}
    \caption{Results of various components of RR2QC combined with Vanilla BERT model on Math Junior dataset.}
    \label{component}
\end{figure}

We conduct ablation experiments on RR2QC's components combined with Vanilla BERT in Math Junior. As shown in Figure \ref{component}, "Base" refers to Vanilla BERT, indicating classification using BERT-Base-Chinese. "TAPT" denotes the pre-training of Vanilla BERT with the MLM task prior to classification. "RCPT" applies our proposed ranking contrastive pre-training task based on TAPT. "CCL" implements Class Center Learning on top of "Base", while "RR" involves reranking the label sequence retrieved by Base using meta-labels.

Figure \ref{component} shows that each component improves upon Vanilla BERT, demonstrating the effectiveness of RR2QC. Notably, RCPT outperforms TAPT in terms of P@1, Macro-F1, and Micro-F1, indicating that learning the fine-grained hierarchical distribution of questions enhances downstream task performance. RR and CCL are both optimized components for the downstream task; while their improvements in Micro-F1 relative to Base are comparable, RR increases P@1 and Macro-F1 by 2.22\% and 2.12\%, respectively, surpassing CCL's improvements of 0.44\% and 0.42\%. This shows that RR places greater emphasis on the recall of long-tail labels, which helps address label imbalance issues.

\subsubsection{Long-tail Test}
We evaluate RR2QC’s performance on imbalanced datasets by comparing its P@1, Micro-F1, and Macro-F1 scores with other methods on head, middle, and tail label subsets of the Math Junior dataset.

As shown in Table \ref{imbanlance test}, RR2QC achieves the highest P@1 (81.81\%) and Micro-F1 (73.52\%) on the head labels, highlighting its strong accuracy on frequent labels. DPT follows closely, with P@1 and Micro-F1 at 80.82\% and 72.63\%, respectively, indicating that prompt tuning and contrastive learning enhance head label recognition.
In the middle label subset, RR2QC again leads with P@1 and Micro-F1 scores of 72.50\% and 64.78\%, demonstrating strong performance on moderately frequent labels. QuesCo follows closely with scores of 70.89\% and 63.34\%, indicating its ranking contrastive pre-training strategy enhances mid-frequency label recognition. For the long-tail labels, RR2QC achieves top scores with P@1 (53.23\%), Micro-F1 (48.65\%), and Macro-F1 (69.04\%), showing high accuracy on rare labels. HPT follows closely, performing effectively on infrequent labels, further validating the effectiveness of prompt tuning. Additionally, traditional models like TextCNN and AttentionXML underperform compared to Transformer-based models across all label subsets, likely due to the absence of a pre-training phase, which limits general knowledge learning. 

Overall, RR2QC demonstrates superior robustness and generalization on long-tail labels, particularly excelling in managing label imbalance. These results confirm the effectiveness of RR2QC in improving performance on imbalanced datasets.

\begin{figure*}[p]
   \scalebox{0.99}{
    \centering
    \includegraphics[width=1.\linewidth]{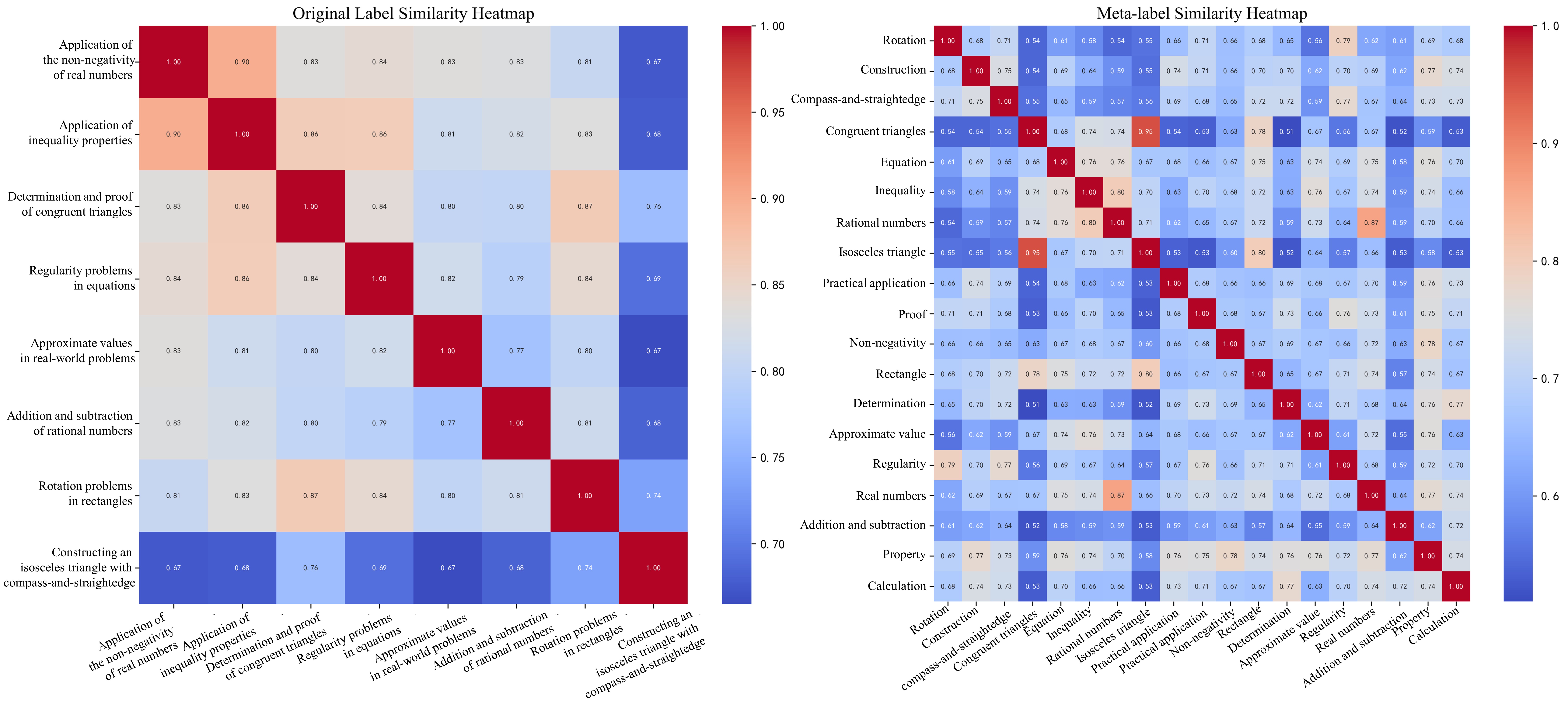}
    }
    \caption{This figure presents similarity heatmaps for 8 randomly sampled labels from Math Junior (left) and their corresponding meta-labels (right). The x-axis and y-axis represent label names translated from Chinese, with higher values indicating a greater degree of similarity. The mean similarity, similarity variance, and proportion of pairs with similarity > 0.8 are 0.7956, 0.0634, and 67.86\% for the left heatmap, while the corresponding values for the right heatmap are 0.6683, 0.0750, and 2.34\%.}
    \label{heatmap}
\end{figure*}
\subsubsection{Comparison between manual and automatic decomposition}\label{Comparison between manual and automatic decomposition}
\begin{table}[h]
    \scalebox{0.9}{
    \centering
    \begin{tabular}{lcccc}
         \hline
         Dataset&L&M&$\overline{\text{L}}$&$\overline{\text{M}}$\\
         \hline
         Math Junior (manual)&721&365&1.11&2.86\\
         Math Junior (auto)&721&587&1.11&3.68\\        
         \hline
         Method&P@1&P@2&Micro-F1&Macro-F1\\
         \hline
         RR2QC (without RR)&77.26&44.03&73.39&63.54\\
         RR2QC (manual)&\textbf{78.36}&\textbf{45.15}&\textbf{74.42}&\textbf{64.03}\\
         RR2QC (auto)&77.63&44.84&73.73&62.96\\
         \hline
    \end{tabular}
    }
    \caption{Comparison results of RR2QC on Math Junior dataset: without RR, with manually decomposed RR, and with automatically decomposed RR. L and M denote the number of labels and meta-labels. $\overline{\text{L}}$ and $\overline{\text{M}}$ denote average number of labels and meta-labels per question.}
    \label{manual and automatic}
\end{table}

In section \ref{manual decomposition}, we engage experts to decompose labels into meta-labels. Since relying on expert-driven decomposition may limit real-world applicability, we automate the process using LLMs. We prompt the LLM to decompose labels into meta-labels. The prompt text $X_{prompt}$ consists of the following three parts:

\textbf{(1) Task description} $X_{desc}$ describes the task of decomposing labels into meta-labels, expressed as:\\
\indent\textit{You are an assistant designed to help decompose input labels into a list of meta-labels. Please ensure that the decomposition minimizes semantic repetition between the meta-labels and maintains a consistent usage of synonyms.}

\textbf{(2) Task Demonstration} $X_{demo}$ consists of manually decomposed pairs for 1\% of the total labels:
\begin{equation*}
    X_{demo} = \{(x^1_{demo}, y^1_{demo})\backslash n,\ldots \backslash n,(x^k_{demo}, y^k_{demo})\}\;,
\end{equation*}
where $x$ represents the label and $y$ is the list of meta-labels. 

\textbf{(3) Input Labels} $X_{labels}$ includes the remaining labels, excluding the demonstrations. 

The prompt text consists of $X_{desc}$, $X_{demo}$, and $X_{labels}$, defined as follows:
\begin{equation*}
    X_{prompt} = \{X_{desc}; \; \backslash n \; X_{demo}; \; \backslash n \; X_{labels};\}\;.
\end{equation*}

We utilize GPT-4o-mini as the prompt backbone in a few-shot environment, iterating multiple times and adopting majority voting—i.e., selecting the most frequently occurring meta-labels from multiple decomposition results as the final output. The experimental results are shown in Table \ref{manual and automatic}. A total of 721 labels are manually decomposed into 365 meta-labels, with an average of 2.86 meta-labels per question. In contrast, the automatic decomposition yields 587 meta-labels, averaging 3.68 meta-labels per question.

In Table \ref{manual and automatic}, both manual and automatic meta-label decompositions improve retrieval reranking (RR) in terms of P@1, P@2, and Micro-F1 compared to RR2QC without RR. Notably, the improvements from expert-driven manual decomposition are significantly higher than those from LLM-based automatic decomposition, demonstrating the effectiveness of meta-label classification and the importance of expert knowledge. 

However, RR with automatic decomposition shows a 0.58\% drop in Macro-F1 compared to RR2QC without RR. This may stem from the excessive number of generated meta-labels (587), some of which lack sufficient independence and synonymy, impacting model performance on imbalanced labels. Therefore, manual review remains essential, and the aforementioned automated decomposition pipeline is available in our GitHub repository.

\subsubsection{Semantic Similarity Analysis of Meta-Labels}

\begin{table}[htbp]
\scalebox{0.70}{
    \centering
    \begin{tabular}{lcccc}
        \hline
        Dataset & Mean Similarity & Similarity Variance & Proportion $>$ 0.8 & Labels \\
        \hline\hline
        MJ & 0.8068 & 0.0614 & 59.59\% & 721 \\
        $\text{MJ}_\text{meta-label}$ & \textbf{0.6464} & \textbf{0.0798} & \textbf{2.02\%} & 365 \\
        $\text{MJ}_\text{meta-label} (\text{auto})$ & 0.6517 & 0.0750 & 2.09\% & 587 \\
        \hline
        MS & 0.8099 & 0.0589 & 61.76\% & 636 \\
        $\text{MS}_\text{meta-label}$ & \textbf{0.6559} & \textbf{0.0772} & \textbf{2.42\%} & 274 \\
        \hline
        PS & 0.7549 & 0.0756 & 29.77\% & 348 \\
        $\text{PS}_\text{meta-label}$ & \textbf{0.6234} & \textbf{0.0822} & \textbf{1.46\%} & 298 \\
        \hline
        DA-20K & 0.7594 & \textbf{0.0799} & 34.16\% & 427 \\
        $\text{DA-20K}_\text{meta-label}$ & \textbf{0.6529} & 0.0797 & \textbf{2.34\%} & 289 \\
        \hline\hline
        WOS & 0.5837 & 0.1169 & 2.73\% & 141 \\
        RCV1-V2 & 0.6333 & 0.0909 & 2.72\% & 103 \\
        \hline
    \end{tabular}}
    \caption{Semantic similarity statistics before and after meta-labeling across different datasets.}
    \label{similarity_metrics}
\end{table}
Meta-labeling aims to decompose semantically similar labels into relatively independent categories, thereby reducing semantic overlap and enhancing the model's recognition ability. To gain an intuitive understanding of how decomposition reduces semantic overlap, we analyze the pairwise similarity between original labels and between meta-labels. This analysis is performed on four educational datasets and two widely used benchmark datasets, including WOS (Web-of-Science) \cite{WOS} and RCV1-V2 \cite{RCV1}.  

We use BERT-Base-Chinese and BERT-Base-Uncased to compute the cosine similarity of label embeddings for Chinese and English datasets. For each dataset, we evaluate the following metrics:

\begin{itemize}
    \item Mean Similarity: Represents the overall similarity level. 
    \item Similarity Variance: Indicates the dispersion of similarity values, reflecting the uniformity of label distribution.  
    \item Proportion of Pairs with Similarity > 0.8: Measures the prevalence of highly similar label pairs.  
\end{itemize}  

A visualization heatmap is provided in Figure \ref{heatmap}, while Table \ref{similarity_metrics} presents detailed similarity metrics across datasets. Notably, meta-labels demonstrate lower mean similarity and a reduced proportion of highly similar pairs. Likewise, the benchmark datasets WOS and RCV1-V2 exhibit both low similarity and a low proportion of highly similar pairs, indicating that a semantically independent and well-structured labeling system effectively reduces redundancy and enhances label differentiation.

\subsubsection{Visualization}
\begin{figure}[htbp]
    \centering
    \subfigure[BERT-TAPT.]
    {
        \begin{minipage}[t]{0.45\linewidth} 
            \raggedright
            \includegraphics[scale=0.25]{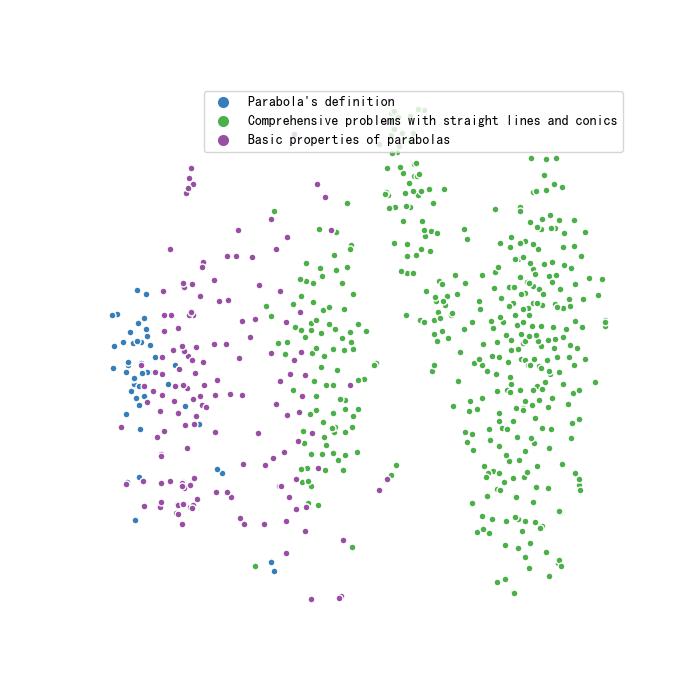}
        \end{minipage}
    }
    \subfigure[BERT-TAPT+CCL.]
    {
        \begin{minipage}[t]{0.45\linewidth}
            \raggedright
            \includegraphics[scale=0.25]{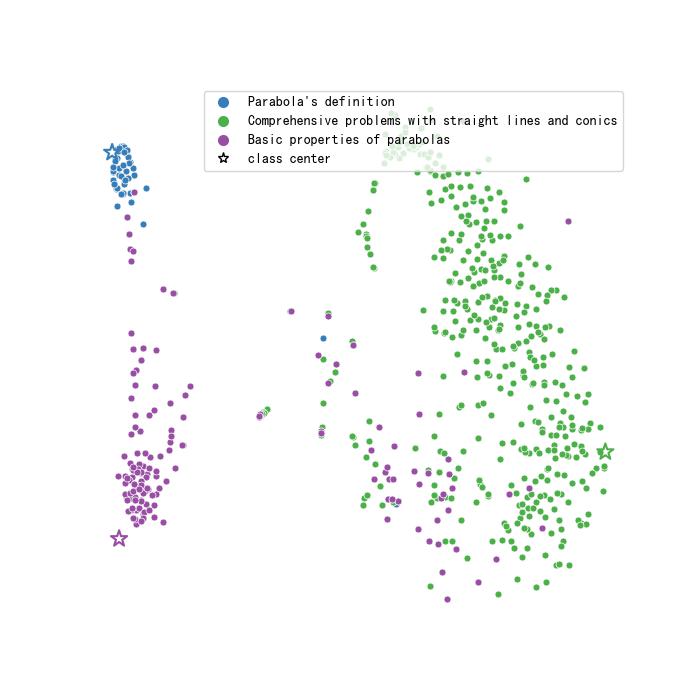}
        \end{minipage}
    }
    \caption{Visualization of questions with three classes.}
    \label{visual}
\end{figure}

Class-Center Learning (CCL) addresses the issue of dense sample distributions for similar labels by attracting samples towards predefined class centers. To illustrate its effectiveness, this study selects three easily confused labels and their corresponding questions from the DA-20K dataset: "Definition of a Parabola," "Basic Properties of Parabolas," and "Comprehensive Problems with Straight Lines and Conics." We use t-SNE \cite{Maaten2008VisualizingDU} to map question features into 2D space. As shown in Figure \ref{visual}, BERT-TAPT struggles to separate the boundaries among the three classes. In contrast, the introduction of CCL enables class centers to act as magnets, grouping corresponding questions and reducing ambiguous samples near decision boundaries.

\subsubsection{Case Study}

\begin{figure}
    \centering
    \includegraphics[width=1.0\linewidth]{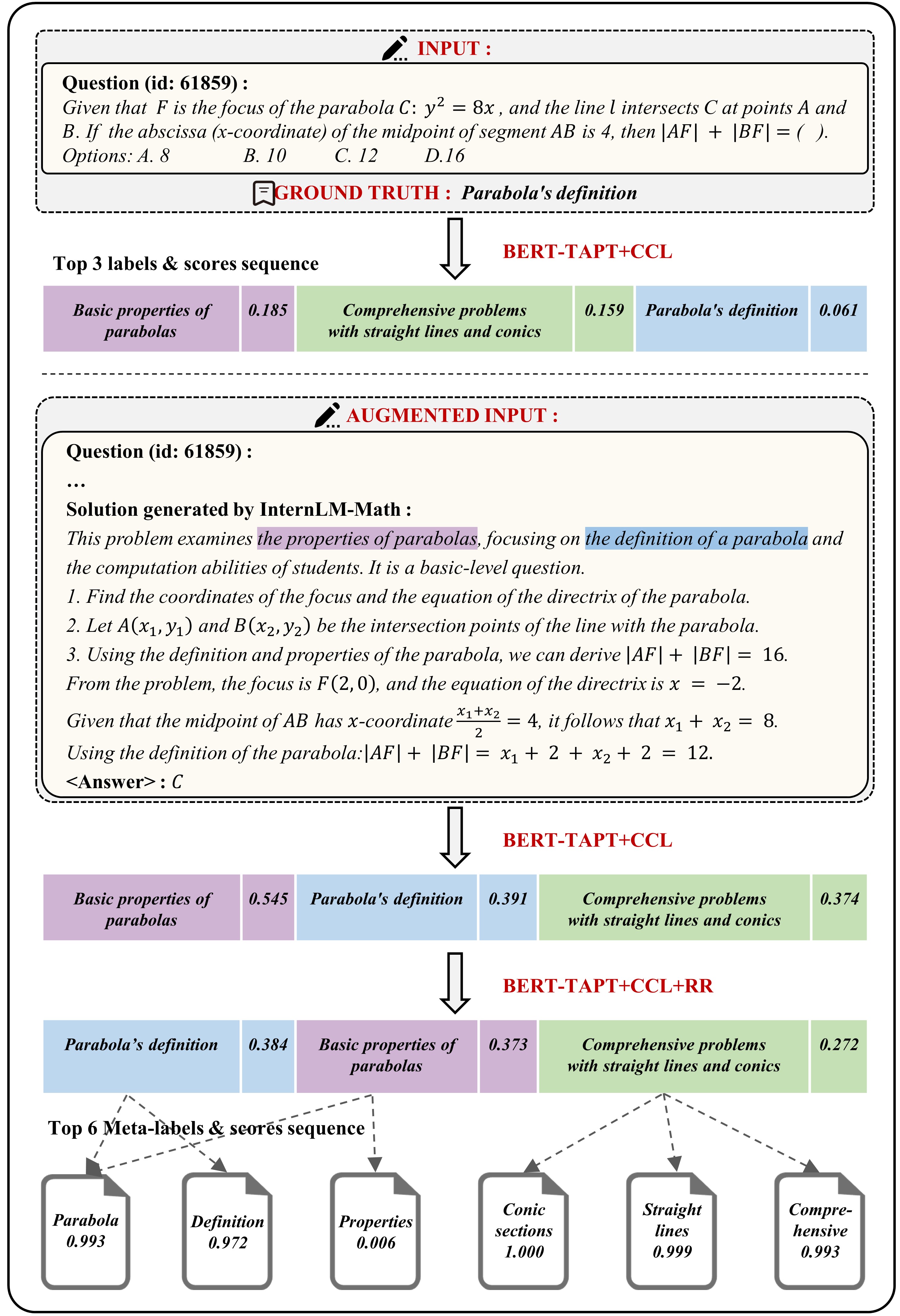}
    \caption{A case study of the ground truth rank changes across RR2QC's components.}
    \label{case study}
\end{figure}

Although CCL markedly improves the model's ability to retrieve the most relevant labels, some questions remain misclassified. This mainly manifests in cases where the ranking of the ground truth improves but does not reach the top position. To address this challenge, this study combines Math LLM-generated solutions to enhance the input and refines label sequences using meta-labels. In Figure \ref{case study}, a representative question from the long-tail label “Parabola's definition” serves as a case study. First, with the aid of CCL, the model elevates the ground truth to 3rd place. Next, the solution generated by InternLM-Math identifies “the properties of parabolas” and “the definition of a parabola” as the relevant knowledge concepts. Then, embedding this solution into the input, while keeping the core model (TAPT+CCL) unchanged, significantly boosts confidence scores, raising the ground truth to 2nd place. Finally, with the help of RR, the label sequence is reranked using Equation \ref{refine}, successfully promoting “Parabola's definition” to 1st place.

\section{Conclusion}\label{Conclusion}
This paper introduces RR2QC, a Retrieval Reranking method for multi-label Question Classification, addressing label imbalance and semantic overlap in educational contexts. By leveraging label semantics, Math LLM-generated solutions, and meta-label refinement, RR2QC significantly enhances classification performance, demonstrating its effectiveness as a robust and versatile solution to improve automated annotation and personalized learning support in online education systems. However, we recognize that relying on expert-annotated meta-labels is not easily scalable. In future work, we plan to automate this process to reduce the expert workload while maintaining quality. Additionally, we aim to explore using LLMs to optimize BERT’s predictions, pushing the limits of traditional classification models rather than simply using data augmentation, as done in this paper.

\section*{CRediT authorship contribution statement}
\textbf{Shi Dong}: Conceptualization, Methodology, Writing – review \& editing, Supervision. \textbf{Xiaobei Niu}: Conceptualization, Methodology, Investigation, Writing – original draft. \textbf{Zhifen Wang}: Conceptualization. \textbf{Rui Zhong}: Conceptualization. \textbf{Mingzhang Zuo}: Funding acquisition, Resources.

\section*{Declaration of competing interest}
The authors declare that they have no known competing financial 
interests or personal relationships that could have appeared to influence 
the work reported in this paper. 
\section*{Data availability}
Links to the data set and our code are given in the manuscript. 
\section*{Acknowledgement}
This work was supported by the National Natural Science Foundation of China (Grant No.61702472, 62002130), Hubei Provincial Natural Science Foundation of China (2022CFB414, 2023AFA020). We also gratefully acknowledge SQUAREMAN Technology Co., Ltd. for providing the MJ, MS, and PS datasets, which are sourced from their exercise collections.


\end{document}